\documentclass[letterpaper]{article} 
\usepackage{aaai2026}  
\usepackage{times}  
\usepackage{helvet}  
\usepackage{courier}  
\usepackage[hyphens]{url}  
\usepackage{graphicx} 
\usepackage{natbib}  
\usepackage{caption} 
\usepackage{algorithm}
\usepackage{algorithmic}
\usepackage{booktabs}
\usepackage{amsmath}
\usepackage{newfloat}
\usepackage{listings}
\DeclareCaptionStyle{ruled}{labelfont=normalfont,labelsep=colon,strut=off} 
\floatstyle{ruled}
\newfloat{listing}{tb}{lst}{}
\floatname{listing}{Listing}

\title{LETToT: Label-Free Evaluation of Large Language Models On Tourism Using Expert Tree-of-Thought}
\author{
    Ruiyan Qi,
    Congding Wen,
    Weibo Zhou,
    Jiwei Li,
    Shangsong Liang,
    Lingbo Li \thanks{Corresponding author. Email: Lingbo.Li.1@warwick.ac.uk}
}

\usepackage{bibentry}

\makeatletter
\renewcommand{\copyright@on}{} 
\makeatother

\begin{document}

\maketitle

\begin{abstract}
Evaluating large language models (LLMs) in specific domain like tourism remains challenging due to the prohibitive cost of annotated benchmarks and persistent issues like hallucinations. 
We propose \textbf{L}abel-Free \textbf{E}valuation of LLM on \textbf{T}ourism using Expert \textbf{T}ree-\textbf{o}f-\textbf{T}hought (LETToT), a framework that leverages expert-derived reasoning structures\textemdash instead of labeled data\textemdash to assess LLMs in tourism.  
First, we iteratively refine and validate hierarchical ToT components through alignment with generic quality dimensions and expert feedback. 
Results demonstrate the effectiveness of our systematically optimized expert ToT with 4.99-14.15\% relative quality gains over baselines.
Second, we apply LETToT's optimized expert ToT to evaluate models of varying scales (32B-671B parameters), revealing: 
(1) Scaling laws persist in specialized domains (DeepSeek-V3 leads), yet reasoning-enhanced smaller models (e.g., DeepSeek-R1-Distill-Llama-70B) close this gap; 
(2) For sub-72B models, explicit reasoning architectures outperform counterparts in accuracy and conciseness ($p<0.05$). 
Our work establishes a scalable, label-free paradigm for domain-specific LLM evaluation, offering a robust alternative to conventional annotated benchmarks.
\end{abstract}

\section{Introduction}
\label{sec:introduction}

Evaluating large language models (LLMs) in domain-specific applications, such as tourism question-answering (QA)~\cite{contractor2019large, yang2020research, wei2024tourllm}, presents significant challenges. 
Traditional evaluation methods~\cite{zhang2023tourism, wei2024tourllm} often depend on costly annotated benchmarks, which are particularly prohibitive in specialized domains like tourism. 
Moreover, LLMs frequently encounter issues such as hallucinations, generating plausible but incorrect information that undermines their reliability. 
These challenges are amplified by the distinct nature of tourism QA, which focuses on practical, travel-related queries requiring real-time data access (e.g., flight statuses, hotel availability) and personalized recommendations based on user preferences~\cite{wang2013case, contractor2019large, martinez2021influence}. 
In contrast, traditional QA typically addresses broader, knowledge-oriented topics in structured settings, relying on static content.

\begin{figure*}[ht]
    \centering
    \includegraphics[width=\textwidth]{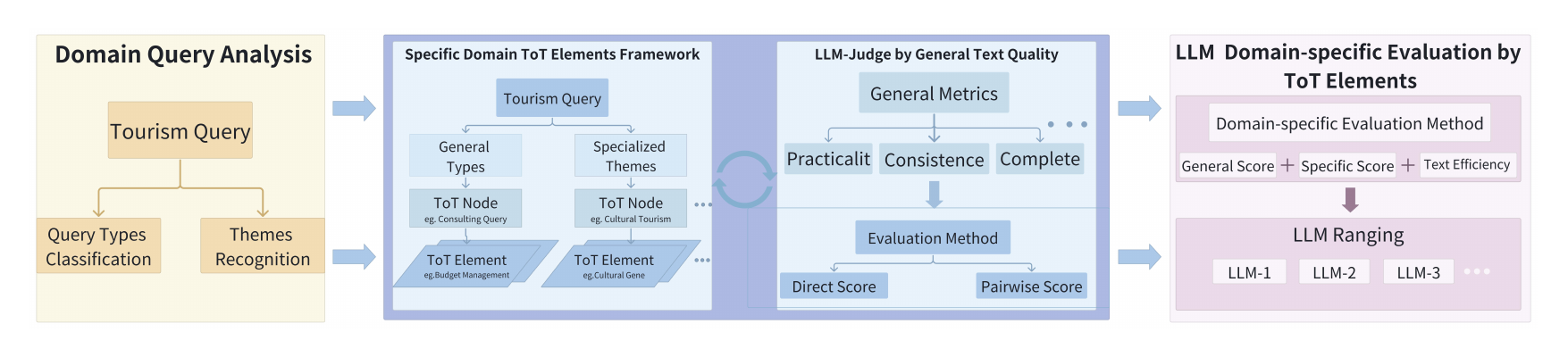}
    \captionsetup{font=scriptsize}
    \caption{Label-Free LLM Evaluation on Tourism using Expert Tree-of-Thought (LETToT) Framework. The LETToT framework integrates three components: (1) \textbf{Domain Query Analysis} (yellow), taxonomically categorizing tourism queries; (2) \textbf{ToT framework} (blue), enabling multi-dimensional LLM-Judge scoring (7 evaluation axes) via direct and pairwise comparative methods; and (3) \textbf{Domain-Specific Evaluation} (purple), benchmarking LLM performance using tailored tourism metrics. }
    \vspace{-1em}
    \label{fig:Label-Free Evaluation of Tourism-Specific LLMs via Expert Tree-of-Thought (LETToT) Framework Queries}
\end{figure*}

To tackle these issues, we propose \textbf{L}abel-Free \textbf{E}valuation of LLM on \textbf{T}ourism using Expert \textbf{T}ree-\textbf{o}f-\textbf{T}hought (LETToT), a novel framework that leverages expert-derived reasoning structures to assess LLMs without the need for labeled data (Figure~\ref{fig:Label-Free Evaluation of Tourism-Specific LLMs via Expert Tree-of-Thought (LETToT) Framework Queries}). 
LETToT is tailored for tourism QA, where queries demand structured reasoning and the integration of user preferences and reasoning to produce coherent travel plans~\cite{ren2024llm, xie2024travelplanner}. 
Despite their potential, LLMs face persistent challenges in tourism, including suboptimal itineraries due to overlooked geographical or user-specific factors, thematic misalignments, and factual inaccuracies (e.g., incorrect operating hours)~\cite{liang2022holistic, zhao2024kgcot, lyu2024crudrag}. 
These issues span seven generic quality dimensions: thematic relevance, context appropriateness, logical coherence, creativity, accuracy, completeness, and practicality. 
Current evaluation methods, such as binary checks, often fail to capture this multidimensional nature~\cite{mizumoto2023aies, xu2025llmreason}.

LETToT addresses these limitations through a two-stage label-free evaluation framework. 
In the first stage, we iteratively validate hierarchical ToT components by aligning them with generic quality dimensions and expert feedback, by prompting LLM with expert-derived ToT then cross-validating the responses with LLM-judge.
Thus we present a systematic approach to discover and validate optimal prompts for tourism QA, and an interpretable grading system optimized via Analytic Hierarchy Process (AHP)-weighted scoring. The final optimized prompts achieve significant improvements in response quality, ranging from 4.99\% to 14.15\% over baseline prompts. 

In the second stage, we use the optimized ToT components obtained from the previous stage as guidelines for label-free LLM evaluation, using a rule-based verifiable reward formula based on coverage of components and text efficiency.

Five open source LLMs with parameter counts ranging from 32B to 671B are selected for experiments. Findings indicate that scaling laws persist in specialized domains, with larger models like DeepSeek-V3 leading in overall performance. 
However, smaller models with enhanced reasoning capabilities, such as DeepSeek-R1-Distill-Llama-70B, can effectively close this performance gap. 
Specifically, for models under 72 billion parameters, those with explicit reasoning capabilities significantly outperform their non-reasoning counterparts in accuracy and conciseness ($p<0.05$), differing from results yielded by generic benchmark~\cite{openllmleaderboard2025}.

These findings underscore the persistence of scaling laws in specialized domains while highlighting the potential of reasoning-enhanced architectures to improve performance in smaller models.

The key contributions of this research are:
\begin{itemize}
  \setlength{\itemsep}{0pt}   
  \setlength{\parskip}{0pt}   
  \setlength{\topsep}{0pt}    
  \setlength{\partopsep}{0pt} 
  
    \item \textbf{Introduction of the Replicable Evaluation Framework: LETToT}: 
    A label-free evaluation method that leverages expert-derived reasoning structures for assessing LLMs in domain-specific applications.
    LETToT provides a scalable, label-free paradigm for domain-specific LLM evaluation, combining domain expertise with general content quality assessment.
    \item \textbf{Demonstration of Prompt Optimization Effectiveness}:
     By optimizing prompts with LETToT's expert ToT, we achieved significant improvements in response quality across multiple dimensions, including thematic relevance (\textbf{+14.15\%}), Context Appropriateness (\textbf{+13.85\%}), and creativity (\textbf{+13.50\%}).
    \item \textbf{Insight into Scaling Laws and Reasoning Capabilities}: 
    Our experiments revealed that while larger models lead in overall performance, smaller models with explicit reasoning architectures can close the performance gap, especially for sub-72B parameter models.
    \item \textbf{Valuable Insights into LLM Capabilities in Tourism QA}: 
    The study offers a comprehensive understanding of how different LLMs perform in tourism, paving the way for future evaluations in other specialized domains.
\end{itemize}

\section{Related Work}
\label{sec:background}

\subsection{LLMs in Tourism QA}
LLMs show significant potential in tourism QA, excelling in multi-turn dialogues, knowledge retrieval, and personalized recommendations through pre-training and fine-tuning~\cite{Kumar2024large, Xia2024leveraging}. For example, Bactrainus enhances complex, multi-hop reasoning in tourism QA~\cite{barati2025bactrainus}. However, challenges like hallucination\textemdash producing plausible but incorrect information\textemdash persist, especially in complex scenarios ~\cite{Kumar2024large, zhao2024kgcot}. LLMs also struggle with novel or cross-domain tasks despite advances in fine-tuning and prompt engineering~\cite{wan2024derailer, yue2024dots}. Integrating external knowledge bases, such as tourism databases, remains challenging~\cite{yue2024dots}.

Recent methods mitigate these issues via advanced reasoning frameworks. RoT~\cite{hui2024rot} uses search tree experiences to improve multi-step QA, while AgentCOT enhances controllability and interpretability through evidence-based, multi-turn generation for justified tourism recommendations~\cite{liang2024textualized}. These advancements highlight LLMs' potential in tourism QA but emphasize the need for better external knowledge integration, dynamic reasoning, and domain-specific fine-tuning~\cite{zhao2024kgcot, yue2024dots, liang2024textualized}.

\subsection{Tree of Thought in LLM QA Systems}
ToT framework enhances LLM reasoning for complex tasks via tree-based search with multi-path exploration ~\cite{yao2023tree}. Extending chain-of-thought reasoning, ToT optimizes problem-solving through lookahead and backtracking~\cite{yao2023tree}. To address local uncertainty, Mo et al. (2023) introduced the Tree of Uncertain Thought (TouT), using probabilistic evaluation to reduce reasoning biases~\cite{mo2023tree}. Gao et al. (2024) improved ToT with Meta-Reasoning Prompting (MRP), dynamically selecting strategies to boost accuracy by 12.3\% in mathematical and coding tasks~\cite{gao2024meta}. Wang et al. (2024) proposed the SEED framework, speeding up ToT by 3.8Ã— while retaining 97\% success rates~\cite{wang2024seed}. While ToT enhances LLM decision-making, its use in tourism QA is underexplored. This study investigates a systematic way to transform domain expert knowledge into optimal ToT prompts.

\subsection{Evaluation of LLM-Generated Tourism Responses}
Evaluating LLM-generated responses in tourism QA is challenging due to factual inaccuracies, cultural insensitivity, and the need for real-time updates. Prior work, such as TourLLM, uses tourism knowledge graphs to enhance recommendations and itinerary planning~\cite{wei2024tourllm}. Existing evaluations often rely on manual annotation, assessing information completeness, logical coherence, and cultural adaptability~\cite{yang2020research}. This study proposes a seven-dimensional evaluation framework to overcome traditional metric limitations, establishing a robust, domain-specific benchmark for LLM performance in tourism QA.

\section{LETToT Framework}
\label{sec:method}
\subsection{Domain Query Analysis}
This study analyzes real-world tourism QA data to categorize inquiries by travel phase and intent, employing a taxonomic approach with empirical induction and data-driven validation. Queries are classified into three types covering the travel lifecycle, supported by established tourism behavior research~\cite{kang2020how, nautiyal2023use}:

\begin{itemize}
  \setlength{\itemsep}{0pt}   
  \setlength{\parskip}{0pt}   
  \setlength{\topsep}{0pt}    
  \setlength{\partopsep}{0pt} 
  
    \item \textbf{Planning}: Focuses on logistics, e.g., `Plan a 3-day culinary itinerary from London to Paris with a Â£1900 budget.'~\cite{kang2020how}
    \item \textbf{Pre-trip consultation}: Seeks destination details, e.g., `What reservations are needed for a Paris trip next month?'~\cite{nautiyal2023use}
    \item \textbf{On-trip guidance}: Provides real-time recommendations, e.g., `Identify must-see exhibition halls in the Louvre.'~\cite{kang2020how}
\end{itemize}

Iterative analysis identified 11 tourism themes: \textit{Cultural, Natural, Hot Spring, Leisure Resorts, Winter Sports, Island, Religious, Urban, Theme Park, Family \& Educational}, and \textit{Wellness}. 
These were derived by mapping tourism products to query patterns, ensuring comprehensive coverage~\cite{zhang2023tourism, maria2016classification}. 
An element-based scoring mechanism enhances response objectivity by quantifying theme-specific characteristics, aligned with real-world travel scenarios~\cite{xiang2011travel}.

\subsection{Iterative Expert ToT Validation and Refinement}

The LETToT's Expert ToT elements integrate the three question types (planning-oriented, pre-trip consultation, and on-trip guidance) with the 11 tourism themes. By embedding this hierarchical structure in prompt engineering, we enhance the quality of LLM-generated responses. The framework's modular design allows for seamless adaptation to diverse tourism contexts through targeted component substitution, ensuring both comprehensive coverage and domain specialization.

\begin{figure}[t]
    \centering
    \includegraphics[width=1\linewidth]{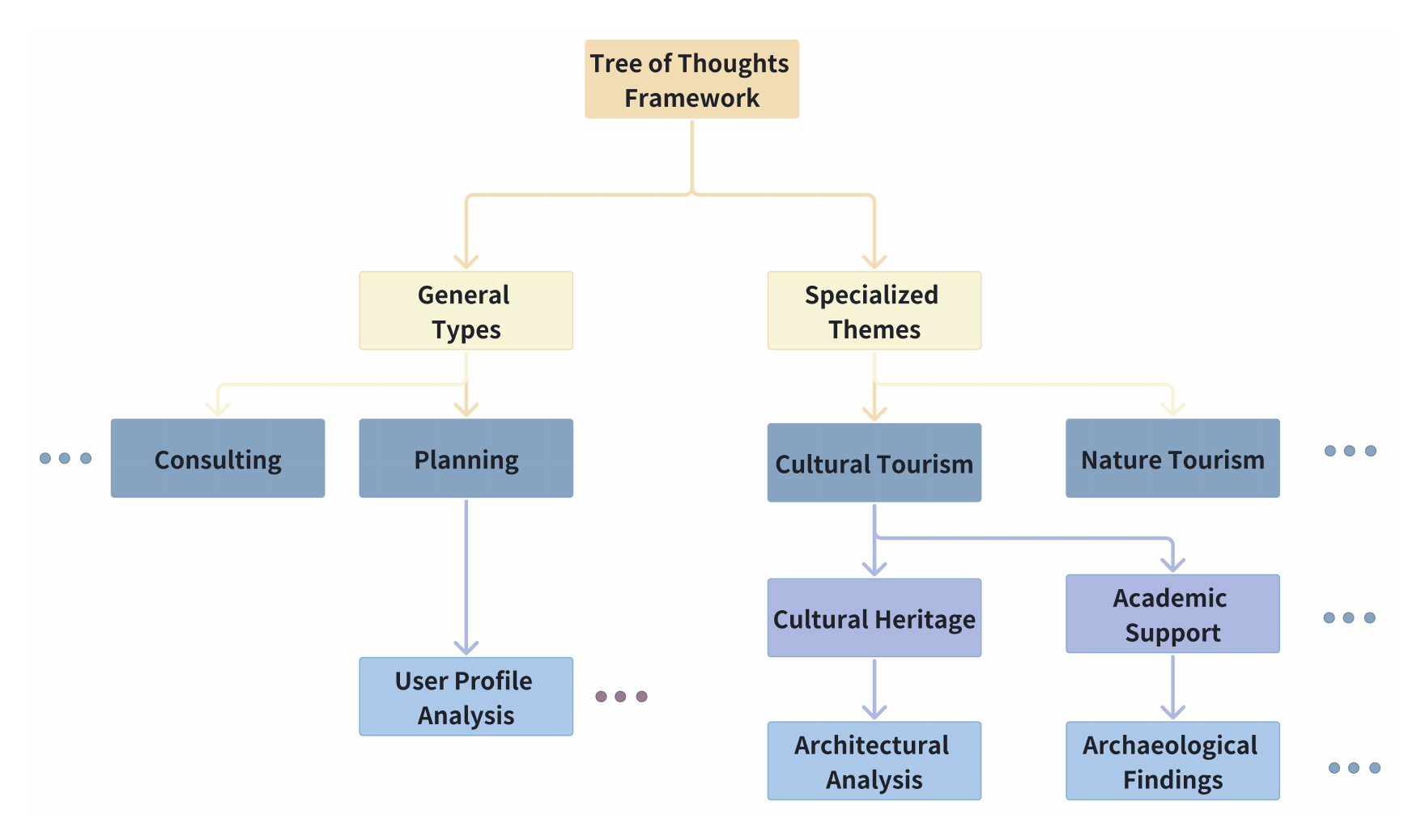}
    \captionsetup{font=scriptsize}
    \caption{LETToT's Expert ToT Framework Structure. The framework employs a dual-layer taxonomy, stratifying general user query types (planning, consulting, guidance)  and 11 tourism subcategories (e.g., cultural tourism) with defined POIs. This design enhances LLM performance via expert-guided prompt optimization.}
    \label{fig:The ToT framework}
\end{figure}

The optimized prompt guided by expert ToT ensures that LLMs generate comprehensive responses with relevant points of interests (POI). For example, a query about a cultural tourism itinerary triggers responses that prioritize relevant POIs, such as historical landmarks or cultural festivals, tailored to the user's travel phase.
Detailed methodologies are presented in Figure~\ref{fig:The ToT framework}.

LETToT's expert ToT framework was refined through iterative validation using LLM-judge, applied to 3,240 tourism-related questions across three travel phases (planning, pre-trip consultation, on-trip guidance) and 11 tourism themes. 

To address potential preference bias in LLM-judge and ensure consistency, two scoring mechanism are employed and cross-validated between each other. Pair-wise score are used as primary metrics since we mainly want to capture the relative performance between optimization and baseline.

\paragraph{Direct Scoring}
The LLM-judge systematically rated response quality using seven generic content quality dimensions: 
Thematic Relevance (\textbf{Rel}), Contextual Adaptability (\textbf{Cxt}), Logical Coherence (\textbf{Log}), Creativity (\textbf{Cr}), Accuracy (\textbf{Acc}), Completeness (\textbf{Comp}), and Practicality (\textbf{Prac}). 
These dimensions, abbreviated as \textit{\textbf{Rel}}-“\textit{\textbf{Prac}}, formed a structured rubric to quantify performance across domain-specific and general content generation criteria (Table \ref{tab:evaluation_metrics}).
Abbreviations are used consistently throughout this work to streamline analysis.
Three independent annotators scored responses on a 1-7 Likert scale (1: extremely poor; 7: excellent), aggregated to ensure inter-rater reliability (Table \ref{tab:evaluation_metrics}). 

\paragraph{Pairwise Scoring}
AHP, a multi-criteria decision analysis method, decomposes complex evaluations into a hierarchical structure of sub-problems, integrating quantitative and qualitative analyses through weighted allocations. In this study, AHP conducted pairwise comparisons between original and optimized response, also across the above mentioned seven dimensions.

\subsection{Domain-Specific Evaluation with LETToT}
This subsection details the second stage of LETToT, leveraging refined ToT elements from the first stage. 
In the evaluation stage, LLMs are prompted with simple query without prior expert knowledge, so as to assess the inherent suitability of LLM for specific domain. 
The evaluation then is designed by systematically identifying and scoring expert ToT elements in the model response, as fine-grained ToT elements make it easily verifiable.

The elements are organized hierarchically, with general elements ensuring broad coverage across query types and specific elements capturing nuanced details relevant to particular tourism themes. This dual structure enables the framework to assess both the comprehensiveness and depth of the content.

\paragraph{Base Score ($S_{\text{base}}$)}
The base score evaluates the coverage of general tourism elements across three categories: planning (e.g., budget management), consultation (e.g., risk assessment), and guidance (e.g., route optimization). It is computed as:

\vspace{-0.5em}
\begin{equation}
\setlength{\abovedisplayskip}{2pt}
\setlength{\belowdisplayskip}{2pt}
\footnotesize
S_{\text{base}} = \sum_{i \in \{P, C, G\}} C_i,
\label{eq:base_score}
\end{equation}
\vspace{-0.5em}

\noindent where \(P\), \(C\), and \(G\) represent the planning, consultation, and guidance categories, respectively, and \(C_i\) is the score for category \(i\), ranging from 0 to 12, evaluated based on predefined criteria for general tourism elements. This score captures the breadth of coverage across essential tourism aspects.

\paragraph{Efficiency Factor ($F_{\text{eff}}$)}
The efficiency factor quantifies the information density of the text using a logistic function. 
It is defined as:

\vspace{-0.5em}
\begin{equation}
\setlength{\abovedisplayskip}{2pt}
\setlength{\belowdisplayskip}{2pt}
\footnotesize
F_{\text{eff}} = \frac{1}{1 + e^{-\frac{N}{L}}},
\label{eq:efficiency_factor}
\end{equation}
\vspace{-0.5em}

\noindent where \(N\) is the total number of elements covered (combining general and specific elements, computed as \(N = \sum C_i + \sum S_j\), with \(S_j\) as the score for each specific element \(j\) in the tourism theme), and \(L\) is the text length in characters. The logistic function, which distinguishes concise texts from verbose ones, leverages the text element density (\(N/L\)) to quantify the concentration of covered elements per character, with higher values indicating greater information efficiency.

\paragraph{Comprehensive Scoring Formula}

To evaluate the quality of tourism-related text, we propose a comprehensive and hierarchical scoring formula that integrates coverage breadth, depth, and information efficiency. The scoring process computed base scores (general tourism elements), specialized scores (theme-specific elements), and efficiency factors (information density). Composite scores were ranked and summarized statistically (mean, max, min, standard deviation). Evaluation protocols are in Tables \ref{tab:elements_framework} and \ref{tab:framework} of Appendix. The updated formula, incorporating weights \(\alpha\) and \(\beta\) for flexibility across different scenarios, is defined as follows:
\begin{equation}
\label{eq:total_score}
   S_{\text{total}} = \left( \alpha S_{\text{base}} + \beta S_{\text{specific}} \right) \cdot F_{\text{eff}} \,,
\end{equation}

\noindent where \(S_{\text{total}}\) is the comprehensive score quantifying overall content quality (rounded to two decimal places for precision), \(S_{\text{base}}\) is the base score assessing the coverage of general tourism elements, \(S_{\text{specific}}\) is the specific score evaluating the coverage of elements unique to a given tourism theme, and \(F_{\text{eff}}\) is the efficiency factor measuring the information density of the text. The weights \(\alpha\) and \(\beta\) are set to 1 by default, allowing equal contribution of \(S_{\text{base}}\) and \(S_{\text{specific}}\), but can be adjusted based on the evaluation context to prioritize either general or specific elements.

\begin{table}[t]
    \centering
    \tiny
    \setlength{\tabcolsep}{3pt}
    \begin{tabular}{lcccc}
        \toprule
        \textbf{Model} & \textbf{Abbreviation} & \textbf{Deployment} & \textbf{Reasoning} & \textbf{Quant} \\
        \midrule
        Qwen2.5-32B-Instruct & Qwen-32B & Local & No & Q4-K-M \\
        Qwen2.5-72B-Instruct & Qwen-72B & Local & No & Q4-K-M \\
        DeepSeek-R1-Distill-Qwen-32B & DS-32B & Local & Yes & Q4-K-M \\
        DeepSeek-R1-Distill-Llama-70B & DS-70B & Local & Yes & Q4-K-M \\
        DeepSeek-V3 & DS-V3 & API & No & N/A \\
        \bottomrule
    \end{tabular}
    \caption{Specifications of Evaluated LLMs.}
    \label{tab:five_models_enhanced}
\end{table}

\section{Experimental Design}

\subsection{Data Preparation}\label{sec:data_preparation}  

A dataset comprising 3,240 records was systematically curated from two primary sources: 
outputs generated by the five open-source LLMs listed in Table \ref{tab:five_models_enhanced} and web-sourced texts extracted from travel forums, blogs, and social media platforms.
The corpus was designed with a 60:40 ratio of LLM-generated to web-sourced content to ensure diversity. 
Pre-processing involved removing non-textual elements, filtering near-duplicates (Levenshtein similarity \textgreater95\%, verified manually), and standardizing text to UTF-8 encoding with uniform casing. 
Data quality was assured through stratified sampling across travel categories (leisure, business, adventure) and manual validation, achieving 95\% annotation agreement.  
The dataset is not available for public release or open-source distribution as it comprises sensitive internal data governed by institutional data governance protocols designed to safeguard proprietary information.

\subsection{Experimental Controls and Optimizations}\label{sec:experimental_controls}

To ensure robust evaluation, a composite scoring formula was developed and validated to distinguish content quality across model outputs, benchmarked against established performance rankings.
Data integrity was maintained by excluding invalid records and implementing computational safeguards to stabilize element identification and scoring processes. 
A standardized logistic function was applied to compute an efficiency factor, fine-tuned through empirical analysis, guaranteeing consistent and reliable assessments.  

\subsection{Evaluation of LLMs}\label{sec:evaluation_llms}

The five open-source LLMs listed in Table \ref{tab:five_models_enhanced} were evaluated under controlled conditions for their performance in tourism QA. 
These models were selected based on their demonstrated capabilities in low-parameter code generation and domain-specific text production, making them suitable for handling varied text structures and travel-related terminology. 
The evaluation setup was designed to ensure reproducibility and accessibility, providing a fair and comprehensive comparison of their effectiveness.

\subsection{Research Questions}

This study investigates two primary research questions to advance the evaluation of LLMs in tourism QA:

\textbf{RQ1: How does incorporating domain-specific expert knowledge into the LLM evaluation pipeline enhance the assessment of model performance in tourism QA?}

Standard evaluation methods often rely on general benchmarks that may overlook the contextual nuances, user preferences, and practical requirements of tourism QA. 
Expert knowledge is critical for assessing the relevance, adaptability, and accuracy of LLM responses in this domain. 
Without exploring how to integrate such knowledge, evaluations may produce misleading results, undermining model selection and development. 
This RQ establishes the foundation for ensuring that evaluations are tailored to the domain, making subsequent research on model optimization meaningful.

\textbf{RQ2: What insights does LETToT provide and how are they compared to traditional supervised benchmarks in assessing LLM performance for tourism QA?}

Supervised benchmarks, while robust, require extensive labeled data, which is resource-intensive to create, particularly in specialized domains like tourism. 
Our label-free framework, LETToT, leverages expert knowledge to evaluate LLMs without labeled data, offering a potentially scalable alternative. 
Validating LETToT against established leaderboards is essential to confirm its reliability and to demonstrate its value as a complementary or alternative approach. 
This RQ ensures that the framework's contributions are rigorously evaluated, justifying its adoption in future research.

\section{Results and Findings}

\textbf{RQ1: How does incorporating domain-specific expert knowledge into the LLM evaluation pipeline enhance the assessment of model performance in tourism QA?}

A controlled experiment compared baseline (raw inputs) and optimized (tourism-expert prompts) responses across 3,240 QA pairs.
Performance was evaluated using the LETToT framework, which integrates domain-specific ToT criteria to assess seven content quality dimensions (\textbf{\textit{Relâ€“Prac}}) via Direct and AHP Pairwise Scoring.

\begin{figure}[t]
    \centering
    \includegraphics[width=1\linewidth]{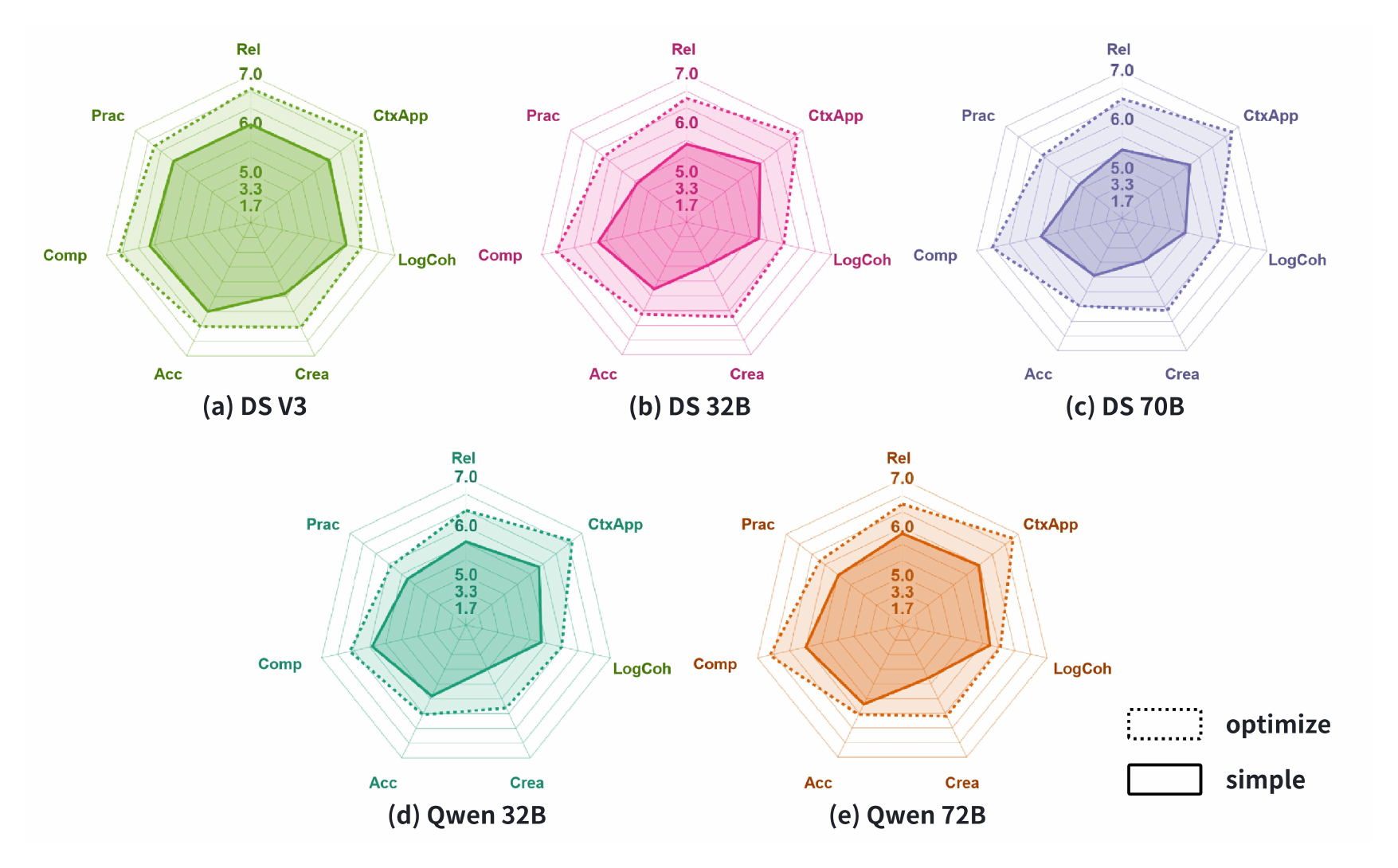}
    \captionsetup{font=scriptsize}
    \caption{Baseline vs. Optimized Prompt Performance Across Quality Dimensions.
    Radar chart compares mean scores (1-7 scale) for baseline and optimized prompts across seven dimensions. 
    Optimized prompts (bold) demonstrate efficacy in enhancing output quality.}
    \label{fig:direct_scoring_radar}
\end{figure}

\begin{figure}[t]
    \centering
    \includegraphics[width=1\linewidth]{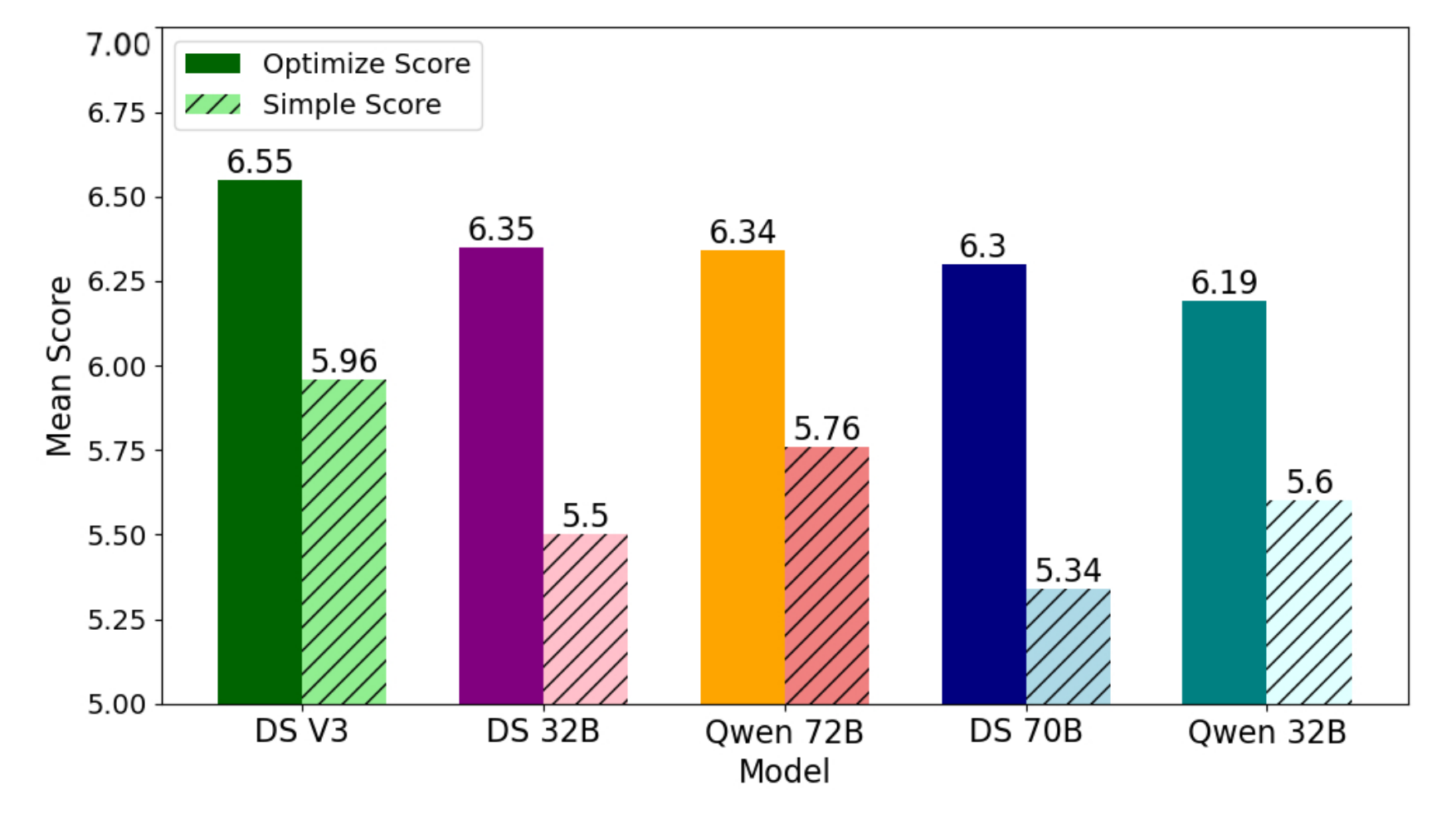}
    \captionsetup{font=scriptsize}
    \caption{Comparative Analysis of Optimized vs. Baseline Prompt Performance.Paired bar plot of mean scores (5.00-7.00, y-axis) across five LLMs (x-axis). Optimized prompts (dark bars) and baseline prompts (light hatched bars) per model highlight performance gains from prompt engineering.}
    \label{fig:Pairwise Analysis Mean Score}
\end{figure}

The expert-guided prompt optimization significantly improved response quality, with gains ranging from +8.23\% (\textbf{\textit{Prac}}) to +14.15\% (\textbf{\textit{Rel}}) (Figure~\ref{fig:direct_scoring_radar}). Reasoning models, in our case DS-32B (+15.15\%) and DS-70B (+17.98\%), exhibited the largest improvements under optimized prompts (Figure~\ref{fig:Pairwise Analysis Mean Score}), surpassing Qwen-72B and Qwen-32B after optimization, demonstrating their enhanced sensitivity to ToT prompting.
Pairwise rankings (Table~\ref{table:Pairwise Ranking}) positioned DS-V3 as the top performer (1st in both baseline/optimized groups).

\begin{table}[t]
\centering
\tiny
\setlength{\tabcolsep}{6.5pt} 
\begin{tabular}{@{}l*{8}{c}@{}}
\toprule
\textbf{Model} & 
\textbf{Rel} & 
\textbf{Cxt} & 
\textbf{Log} & 
\textbf{CR} & 
\textbf{ACC} & 
\textbf{Comp} & 
\textbf{Prac} & 
\textbf{OR} \\
\midrule
DS-32B  & 4,\textbf{2} & \underline{3},\textbf{2} & 4,\underline{3} & 4,\textbf{2} & 4,\textbf{2} & 4,\underline{3} & 4,\textbf{2} & 4,\textbf{2} \\
DS-70B  & 5,4 & 5,4 & 5,4 & 5,\underline{3} & 5,5 & 5,4 & 5,4 & 5,4 \\
DS-V3   & 1,1 & 1,1 & 1,1 & 1,1 & 1,1 & 1,1 & 1,1 & 1,1 \\
Qwen-32B & \underline{3},5 & 4,5 & \underline{3},5 & \underline{3},5 & \underline{3},4 & \underline{3},5 & \underline{3},5 & \underline{3},5 \\
Qwen-72B & \textbf{2},\underline{3} & \textbf{2},\textbf{2} & \textbf{2},\textbf{2} & \textbf{2},4 & \textbf{2},\underline{3} & \textbf{2},\textbf{2} & \textbf{2},\underline{3} & \textbf{2},\underline{3} \\

\bottomrule
\end{tabular}
\captionsetup{font=scriptsize}
\caption{Comparative Ranking of Baseline (S) vs. Optimized (O) Prompts by Quality Dimension.
Rankings (1 = best) formatted as S, O, with lower values indicating superior performance. 
Second- and third-place entries are \textbf{bolded} and \underline{underlined}, respectively. 
OR denotes Overall Rank.}
\label{table:Pairwise Ranking}
\end{table}

The LETToT framework quantified these enhancements through AHP analysis (Table~\ref{tab:average_scores}), revealing DS-32B's dominance in relevance (6.08/7) and DS-70B's strength in contextual adaptability (5.34/7).
This validates LETToT's ability to rigorously evaluate domain-specific LLM performance without manual labels, demonstrating the critical role of expert knowledge in measuring and guiding tourism-focused QA improvements.

\begin{table}[t]
\centering
\scriptsize
\setlength{\tabcolsep}{6.5pt}
\begin{tabular}{@{}lccccccc@{}}
\toprule
Model & Rel. & Cxt. & Log. & Cr. & Acc. & Comp. & Prac. \\
\midrule
Qwen-32B & 4.52 & 4.99 & 4.37 & 3.95 & 4.88 & 5.19 & 5.17 \\ 
Qwen-72B & 4.77 & 5.15 & 4.49 & 4.34 & 4.73 & 5.62 & 5.55 \\
DS-70B & \underline{5.19} & \textbf{5.34} & \textbf{4.98} & \underline{4.90} & \textbf{5.11} & \textbf{5.90} & \textbf{5.77} \\
DS-32B & \textbf{6.08} & \underline{5.27} & \underline{4.92} & \textbf{4.76} & \underline{5.03} & \underline{5.79} & \underline{5.68} \\ 
DS-V3 & 4.88 & 5.25 & 4.50 & 4.44 & 4.72 & 5.55 & 5.49 \\
\bottomrule
\end{tabular}
\captionsetup{font=scriptsize}
\caption{AHP Evaluation of LLMs Across Generic Quality Dimensions.Hierarchical scoring (1-7 scale) for seven dimensions. 
\textbf{Bold} = highest score, \underline{underlined} = second-highest per column. }
\label{tab:average_scores}
\end{table}

\textbf{Answer to RQ1}: 
Incorporating domain-specific expert knowledge via optimized prompts enhances LLM performance in tourism QA by 4.99--14.15\% across seven metrics. with DS-V3 outperforming Qwen-72B, DS-32B/70B, and Qwen-32B in baseline and optimized settings. 
The LETToT framework's effectiveness is validated through pairwise comparisons and AHP-weighted scoring, ensuring robust, domain-tailored assessment.

\textbf{RQ2: What insights does LETToT provide and how are they compared to traditional supervised benchmarks in assessing LLM performance for tourism QA?}

To compare domain-specific LLM performance with established leaderboards, this study evaluates task-specific scores, average performance, text length efficiency, and statistical significance.

\begin{figure}[t]
    \centering
    \includegraphics[width=0.95\linewidth]{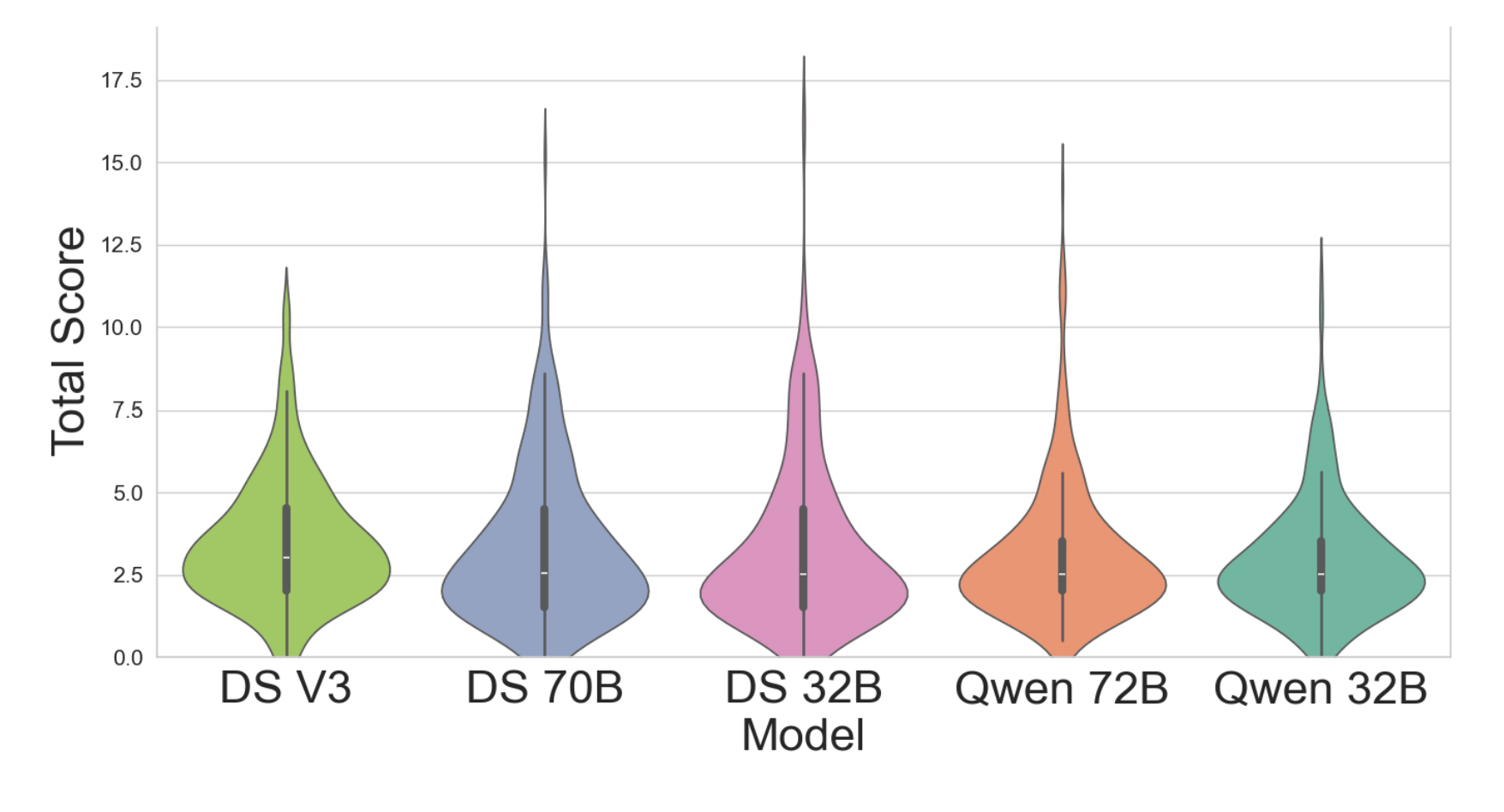}
    \captionsetup{font=scriptsize}
    \caption{Violin Plot of LLM Comprehensive Score Distributions.
    KDE-smoothed distributions of performance scores (y-axis: 0-20) across five LLMs (x-axis). 
    Violin widths reflect probability density at score intervals, while embedded boxplots denote median values and interquartile ranges.}
    \label{fig:violin}
\end{figure}

Violin plots (Figure~\ref{fig:violin}) reveal DS-V3 achieved the highest mean score (3.47 $\pm$ 0.11, 95\% CI [3.27--3.68], IQR = 2.52), followed by DS-70B (3.34, 95\% CI [3.09--3.59], IQR = 3.0) and DS-32B (3.30, 95\% CI [3.03--3.57], IQR = 3.0). 

\begin{figure}[t]
     \centering
     \includegraphics[width=0.95\linewidth]{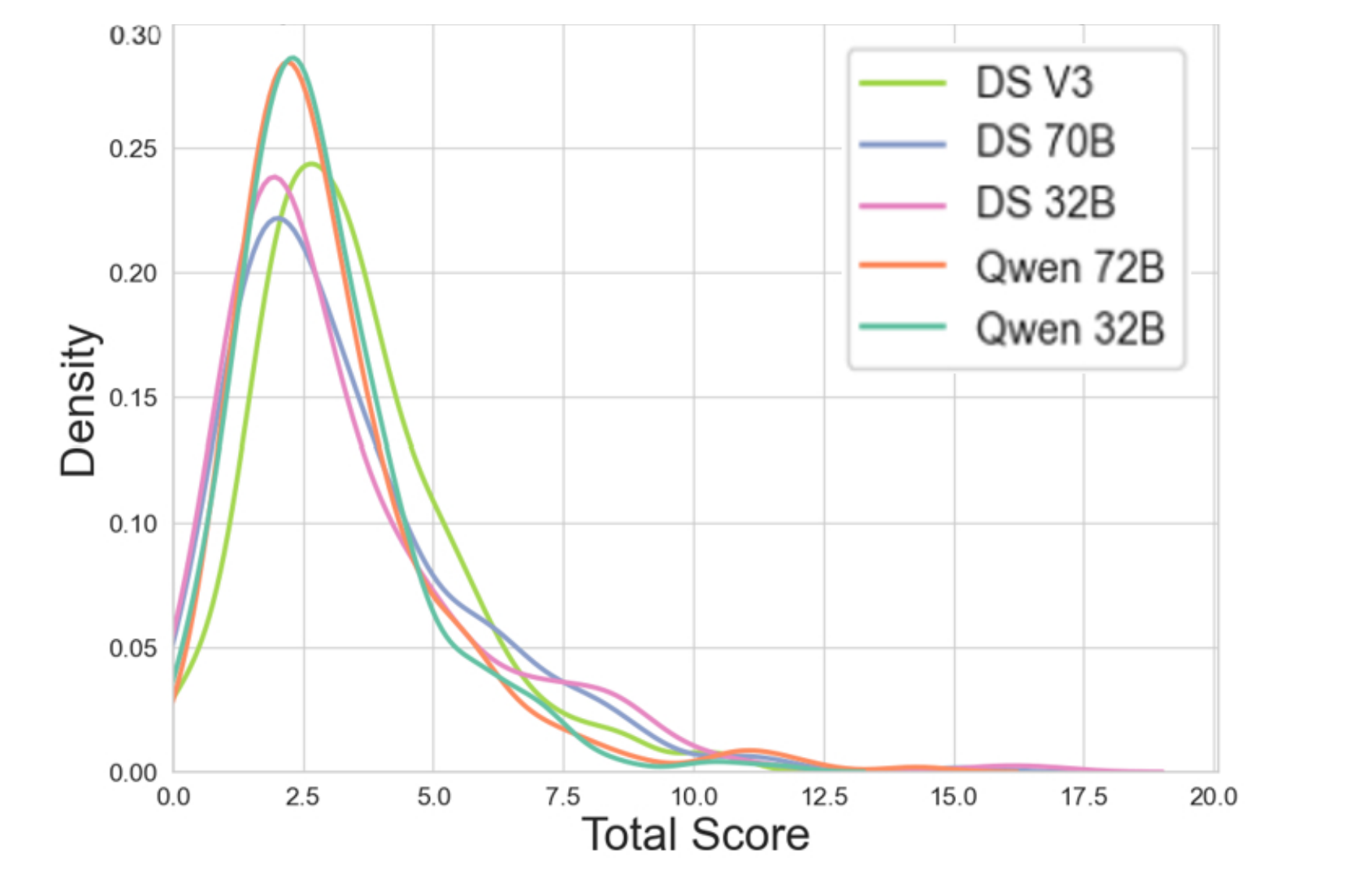}
     \captionsetup{font=scriptsize}
     \caption{Probability Density Distributions of LLM Performance Scores.
     This KDE plot comparing score distributions (range: 0-0, x-axis) across five LLMs. 
     The y-axis quantifies probability density, with colored curves representing individual model distributions. 
     Overlapping density profiles indicate convergent performance characteristics among models.}
     \label{fig:density}
\end{figure}

Qwen-72B (3.11, 95\% CI [2.89--3.33], IQR = 1.53) and Qwen-32B (2.96, 95\% CI [2.77--3.15], IQR = 1.53) scored lower with tighter distributions. 
DS models exhibited wider confidence intervals (average span = 0.38) than Qwen models (0.34), indicating greater variability but superior performance.
Density estimates (Figure~\ref{fig:density}) show that DS-V3 exhibits the highest peak and stable performance with a moderate spread, DS-70B and DS-32B have flatter distributions with wider ranges, while the Qwen models display tighter, left-skewed distributions, indicating limited peak performance. Overall, DS models offer higher scores with greater variability, whereas Qwen models focus on compactness at the cost of peak performance.

\begin{figure}[t]
    \centering
    \includegraphics[width=0.9\linewidth]{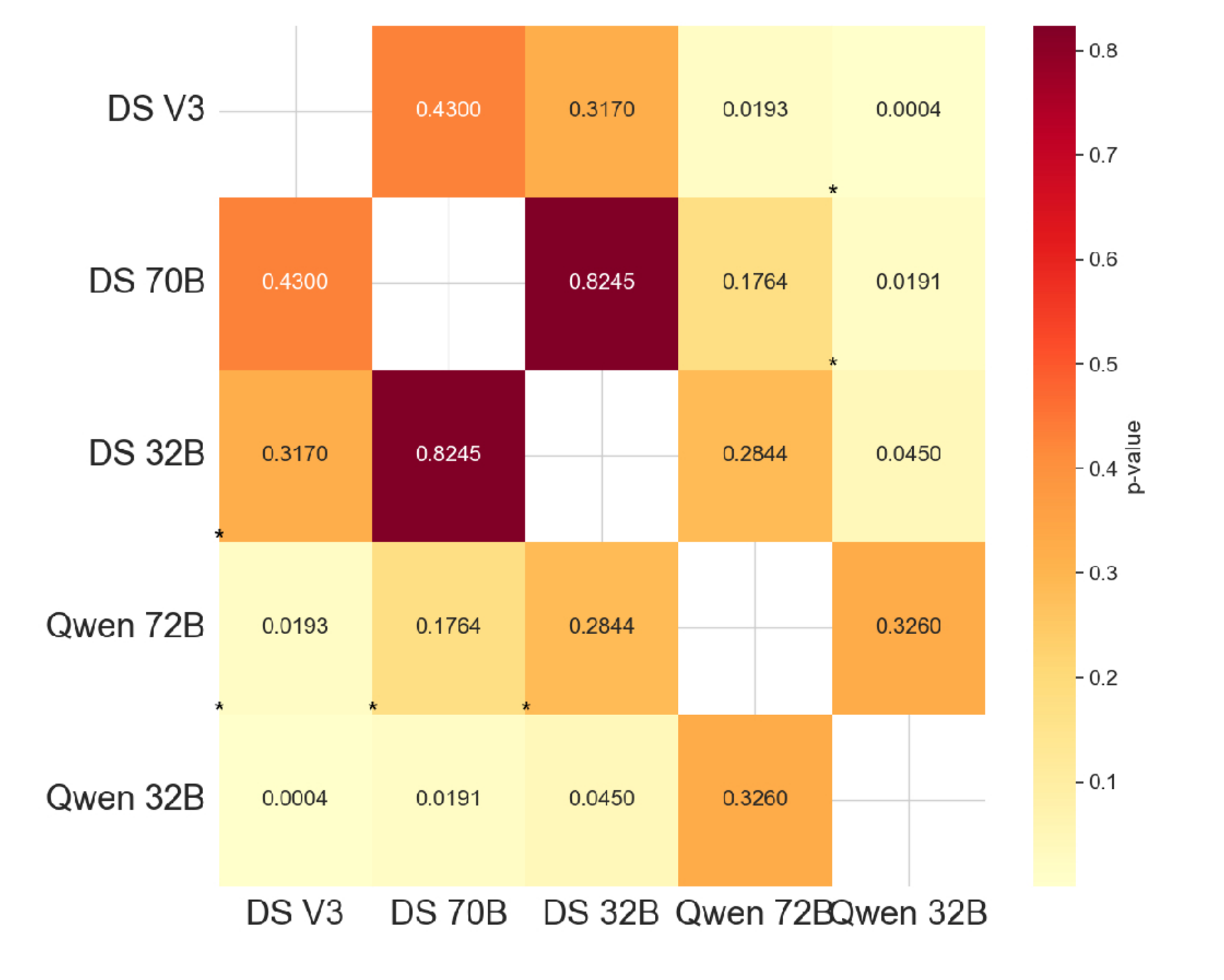}
    \captionsetup{font=scriptsize}
    \caption{Pairwise T-Test P-Value Matrix for LLM Performance Comparisons.
    Heatmap visualization of p-values derived from pairwise t-tests between five evaluated LLMs. 
    Axes represent models, forming a symmetric pairwise comparison matrix. 
    Color intensity scales from light hues (low p-values, e.g., 0.0004, indicating significant differences) to dark red (high p-values, e.g., 0.450, denoting negligible statistical distinctions).}
    \label{fig:heatmap}
\end{figure}

Statistical analysis via a p-value heatmap (Figure~\ref{fig:heatmap}) confirms DS-V3 significantly outperforms Qwen-72B (p = 0.019) and Qwen-32B (p = 0.0004), with DS-70B and DS-32B surpassing Qwen-32B (p = 0.019 and p = 0.045, respectively). 
No significant differences were observed among DS models (p \textgreater 0.32). 
Text length analysis indicates that concise outputs (\textless500 tokens) correlate with higher scores, with longer outputs yielding diminishing returns.

\begin{figure}[t]
    \centering
    \includegraphics[width=0.9\linewidth]{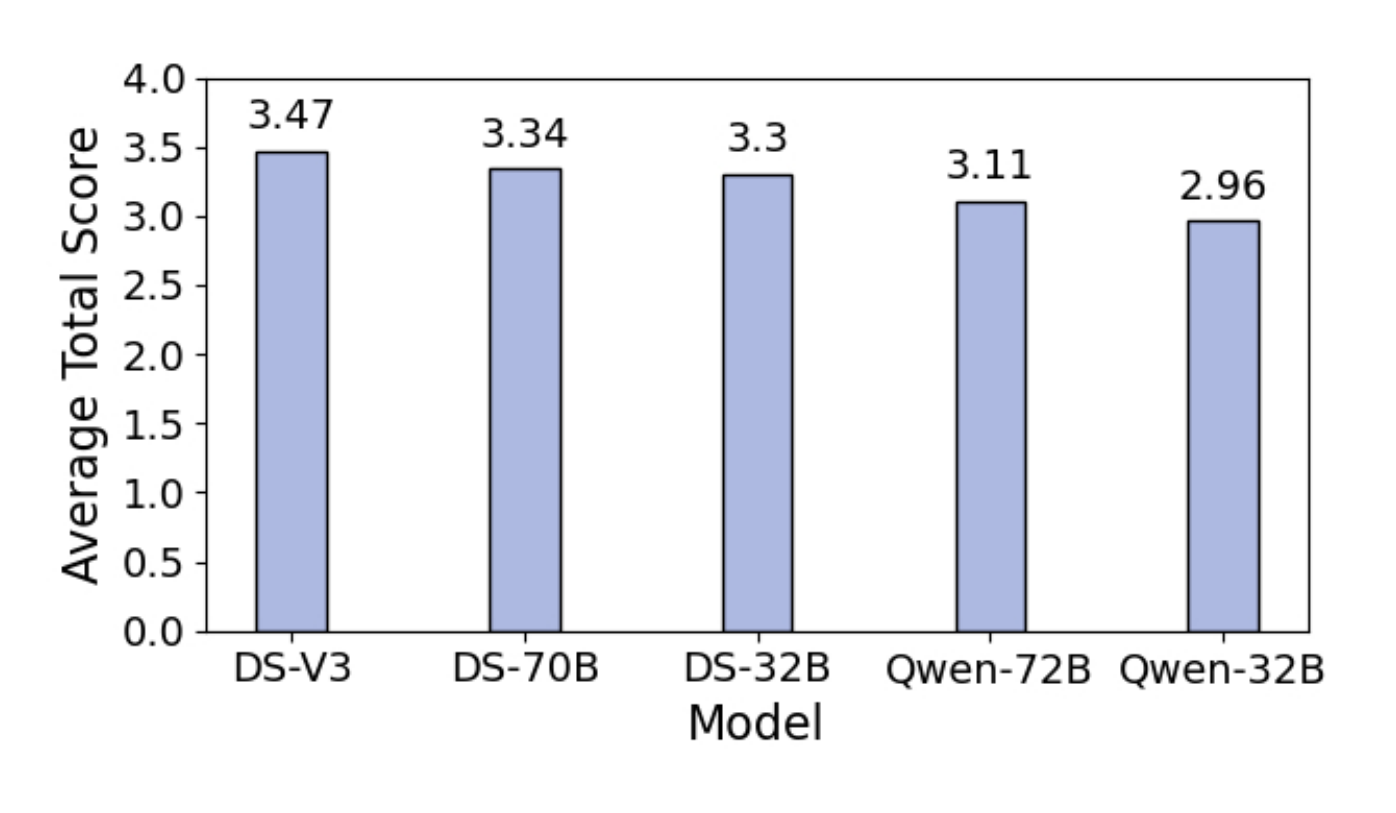}
    \captionsetup{font=scriptsize}
    \caption{Comparative Capability Evaluation of LLMs.
    Bar plot quantifies average performance scores (0-4 scale) across five LLMs. 
    Data labels and bar heights reflect ranking and exact scores, with taller bars indicating superior capability.
    Bars are sorted left-to-right in descending order of average scores.}
    \label{fig:barplot}
\end{figure}

The evaluation results present a clear performance hierarchy (Figure~\ref{fig:barplot}): DS-70B achieved the highest score among listed models (3.34), followed closely by DS-32B (3.30), Qwen-72B (3.11), and Qwen-32B (2.96). 
Notably, within the 32B--72B parameter range, reasoning-enhanced models (DS-70B, DS-32B) significantly outperforms non-reasoning models (Qwen-72B, Qwen-32B) with concise outputs ($p<0.05$), highlighting reasoning's critical role in tourism QA.

Comparing this ranking obtained by LETToT with the Direct Scoring by LLM-judge on generic content qualities in the first stage experiment, we notice that within the 32B--72B parameter range, the relative positions of DeepSeek models and Qwen models are reversed. While Qwen models score higher according to generic LLM-judge, the DeepSeek models with explicit reasoning outperform in LETToT's ranking.

We further compare with popluar generic LLM Leaderboards like HuggingFace's Open LLM Leaderboard~\cite{openllmleaderboard2025}.
Note DS-V3 exceeds the Open LLM Leaderboard's typical range (up to 140 billion parameters) and is not listed.

Looking at the remaining models and their comparable rankings, the same discrepancy persists. 
While by LETToT ranking, DS-70B and DS-32B with reasoning abilities clearly outperform Qwen-72B and Qwen-32B, on Open LLM leaderboard their positions are reversed, by a large margin, where Qwen-72B, Qwen-32B, DS-70B and DS-32B are ranked at 6th, 22nd, 1320th and 2100th respectively.

This discrepancy demonstrate LETToT's ability to provide domain-specific assessment that differs from generic evaluation, while highlighting the importance of reasoning ability in fields requiring expertise knowledge.

\textbf{Answer to RQ2}: 

The LETToT framework provides domain-specific insights into LLM performance for tourism QA, ranking DS-V3 highest, followed by DS-70B, DS-32B, Qwen-72B, and Qwen-32B.
Reasoning-enhanced models (32B--72B) significantly outperform non-reasoning counterparts in accuracy and conciseness (p < 0.05), a nuance less evident in supervised benchmarks like the HuggingFace Open LLM Leaderboard. 
LETToT's label-free approach enhances scalability and captures tourism-specific requirements, offering a robust alternative for specialized evaluation.

\section{Threat to Validity}

\textbf{Internal Validity:} 
Internal validity threats include experimental inconsistencies and subjective evaluation biases. 
Variations in responses across LLMs, such as DeepSeek and Qwen, could arise from differences in model architectures or API configurations.
To mitigate this, we standardized API parameters and conducted five independent runs per experiment, ensuring consistent testing conditions. 
Subjective evaluation biases were addressed using a dual scoring mechanism combining direct and AHP-weighted scoring, with blind scoring by three domain experts followed by consensus discussions to enhance objectivity.

\textbf{External Validity:}
External validity threats stem from the dataset's focus on Chinese urban itineraries and potential cultural gaps in the seven-dimensional metrics (Thematic Relevance, Context Appropriateness, logical coherence, creativity, accuracy, completeness, and practicality). 
To enhance generalizability, we aligned the dataset with UNESCO and United Nations World Tourism Organization (UNWTO) frameworks, ensuring it reflects globally relevant tourism concepts. 
Additionally, we validated the metrics with native experts from five diverse regions to confirm their applicability across different cultural contexts.

\section{Conclusion}\label{sec:conclusion}
Current evaluation LLMs in highly specialized domains like tourism remain constrained by reliance on annotated benchmarks and inadequate domain specificity, failing to assess critical competencies like cultural contextualization or itinerary planning, resulting in increasing discrepancies between domain expert opinions and general assessment.

To address this, we present \textbf{LETToT} (Label-Free Evaluation of LLMs on Tourism via Expert Tree-of-Thought), a domain-specific evaluation framework that eliminates manual annotation costs by leveraging structured reasoning hierarchies derived from tourism expertise.

The LETToT framework offers two key innovations in methodology.

\textbf{Domain-Aware Evaluation Design with Iterative Refinement}:
Tourism queries are taxonomically classified (products/policies/general) and mapped to expert-validated quality dimensions (e.g., Thematic Relevance, Practicality).
ToT hierarchies replaces labeled data, iteratively refined and validated by LLM-judge, enabling granular assessment of LLM outputs. 

\textbf{Label-Free Quality Assessment}:
ToT reasoning chains serve as automated evaluation oracles, rigorously scoring LLM responses via AHP-weighted multi-criteria analysis.

We validated LETToT's effectiveness through a large-scale experiments involving five LLMs and 3,240 QAs. The results yields interesting insights on how LLMs perform in specialized domain: 
(1) Expert knowledge could enhance performance across the board (4.99-4.15\%), while reasoning models show enhanced adaptivity to optimized prompts.
(2) Although reasoning models could be less consistent in performance, they outperform non-reasoning models in aligning with domain expert expectation, which is often overlooked in generic evaluations.

Replacing costly annotations with expert-guided reasoning hierarchies, LETToT establishes a scalable, domain-oriented 
paradigm for evaluating tourism QA systems. The same framework can be extended to other fields that require deep domain expertise.

\clearpage
\bibliography{aaai2026}
\newpage
\begin{onecolumn}
\section{Appendix}\label{app:eval_protocols}

This appendix provides supplementary tables and detailed descriptions not included in the main text. These include the \textbf{Definitions of Scoring Dimensions and Their Scale Interpretations} (Table \ref{tab:evaluation_metrics}), the \textbf{Tourism General ToT Elements Framework} (Table \ref{tab:elements_framework}), and the \textbf{Tourism Characteristic ToT Elements Framework} (Table \ref{tab:framework}). 

The \textbf{Direct Scoring} process employs a structured rubric within the LLM-Judge framework to quantitatively assess LLM performance across seven dimensions. These dimensions evaluate both domain-specific and general content generation criteria. Table \ref{tab:evaluation_metrics} provides precise definitions for each dimension, accompanied by explanations and a 1--7 Likert scale for consistent performance measurement. The scale ranges from 1 (lowest performance) to 7 (highest performance), enabling composite scores that support quantitative comparisons between baseline and optimized LLM outputs.

The \textbf{Comprehensive Scoring Formula}, as introduced in the main text, evaluates content generated by LLMs by integrating multiple indicators. These indicators encompass base scores derived from general tourism elements and specialized scores reflecting theme-specific elements. The general tourism elements are detailed in Table \ref{tab:elements_framework}, which outlines the \textbf{Tourism General ToT Elements Framework}. This framework categorizes elements into three tourism query types: \textbf{Planning}, \textbf{Consulting}, and \textbf{Guiding}, with each category comprising six distinct elements. These elements collectively address the core components of each query type, ensuring a robust evaluation structure.

Additionally, Table \ref{tab:framework} presents the \textbf{Tourism Characteristic ToT Elements Framework}, which includes eleven tourism categories. Each category is defined by three specific elements, and every element is further subdivided into four detailed sub-elements (two in Part 1 and two in Part 2). This hierarchical organization facilitates a thorough analysis of theme-specific components, enhancing the granularity of the scoring process.

\begin{table*}[ht]
\centering
\scriptsize
\resizebox{\textwidth}{!}{
\begin{tabular}{lccc}
\toprule
\textbf{Dimension} & \textbf{Concept Explanation} & \multicolumn{2}{c}{\textbf{Evaluation Scale Meaning}} \\
\cmidrule{3-4}
 &  & \textbf{Lowest (1)} & \textbf{Highest (7)} \\
\midrule
\textbf{\parbox{3cm}{Thematic Relevance (Rel)}} & \raggedbottom Alignment degree between answer content and tourism core themes (scenic spots, culture, history, travel services). & \raggedbottom Completely irrelevant. Off the point. & \raggedbottom Fully adheres to theme, covers all elements. \\ 
\textbf{\parbox{3cm}{Context Appropriateness\newline(Cxt)}} & \raggedbottom Match between answer and user's specific scenario needs (family outings, business trips, cultural study tours). & \raggedbottom Incompatible. Incorrect suggestions. & \raggedbottom Fully adapted (e.g., cultural taboos, LNT principles). \\ 
\textbf{\parbox{3cm}{Logical Coherence (Log)}} & \raggedbottom Information organization clarity, natural transitions between steps/viewpoints, and no contradictions. & \raggedbottom Fragmented content, lacks logic. & \raggedbottom Clear structure, rigorous logic, distinct layers. \\ 
\textbf{\parbox{3cm}{Creativity (Cr)}} & \raggedbottom Novelty in tourism suggestions, avoiding template-based content. & \raggedbottom Useless/impractical suggestions. & \raggedbottom Unique, attractive \& relevant suggestions. \\ 
\textbf{\parbox{3cm}{Accuracy (Acc)}} & \raggedbottom Objective correctness of information (opening hours, ticket prices, history) with authoritative verification. & \raggedbottom Completely wrong/fictional. & \raggedbottom Fully accurate with no errors. \\ 
\textbf{\parbox{3cm}{Completeness (Comp)}} & \raggedbottom Covers all key points in the user's question, supplementing related info when needed (transportation, precautions). & \raggedbottom No valid information. & \raggedbottom Comprehensive coverage with expansions. \\ 
\textbf{\parbox{3cm}{Practicality (Prac)}} & \raggedbottom Operational content guiding user decisions/actions (reservation links, contacts, operation guides). & \raggedbottom No practical value. & \raggedbottom Highly executable, covers main requirements. \\
\bottomrule
\end{tabular}
}
\captionsetup{font=scriptsize}
\caption{Definitions of Scoring Dimensions and Their Scale Interpretations. This table outlines seven dimensions for assessing responses to tourism-related queries, with detailed explanations provided in the \textbf{Concept Explanation} column. The \textbf{Evaluation Scale Meaning} column, subdivided into \textbf{Lowest (1)} and \textbf{Highest (7)}, delineates the interpretive range of the 1--7 Likert scale, where 1 indicates the lowest performance level and 7 indicates the highest.}
\label{tab:evaluation_metrics}
\end{table*}

\begin{table}[h]
\centering
\scriptsize
\begin{tabular}{ll}
\hline
\textbf{Category} & \textbf{Element} \\
\hline
Planning & $\bullet$ Budget Management \\
         & $\bullet$ Safety System \\
         & $\bullet$ Transportation Network \\
         & $\bullet$ User Profile Analysis \\
         & $\bullet$ Technology Application \\
         & $\bullet$ Sustainable Development Strategy \\
\hline
Consulting & $\bullet$ Information Update Timeliness \\
           & $\bullet$ Risk Assessment \\
           & $\bullet$ Dynamic Early Warning \\
           & $\bullet$ Policy Compliance \\
           & $\bullet$ Community Engagement Mechanism \\
           & $\bullet$ Multilingual Support \\
\hline
Guiding & $\bullet$ Route Design \\
        & $\bullet$ Accessibility Facilities \\
        & $\bullet$ Emergency Support \\
        & $\bullet$ Cultural/Ecological Interpretation \\
        & $\bullet$ Interactive Experience \\
        & $\bullet$ Service Response System \\
\hline
\end{tabular}
\captionsetup{font=scriptsize}
\caption{Tourism General ToT Elements Framework. This table presents a Tourism ToT framework for general elements, organized into three categories: \textbf{Planning}, \textbf{Consulting}, and \textbf{Guiding}. Each \textbf{Category} includes six \textbf{Elements}.}
\label{tab:elements_framework}
\end{table}

\begin{table}[t]
\centering
\scriptsize
\begin{tabular}{llll}
\hline
\textbf{Category} & \textbf{Element} & \textbf{Sub-Element (Part 1)} & \textbf{Sub-Element (Part 2)} \\
\hline
Cultural Tourism & Cultural Heritage & Historical Event Correlation & Intangible Cultural Heritage Transmission \\
 &  & Architectural Analysis & Multi-Faith Comparative Display \\
 & Academic Support & Literature Reference System & Multilingual Interpretation Database \\  
 &  & Archaeological Findings Publication & Interdisciplinary Research \\
 & Experience Design & AR/VR Temporal Narrative & Cultural Consumption Product Innovation \\
 &  & Traditional Festival Revitalization & Flow Experience Optimization \\
\hline
Nature Tourism & Ecological Protection & Endangered Species Monitoring & Carbon Sink Enhancement Measures \\
 &  & Habitat Buffer Zone Design & Soil and Water Erosion Control \\
 & Educational System & Biodiversity Interpretation System & Nature Material Creation Workshop \\
 &  & Ecological Restoration Participation & Modular Study Tour Curriculum \\
 & Community Collaboration & Indigenous Employment Training & Traditional Ecological Knowledge Application \\
 &  & Local Supply Chain Management & Disaster Co-Prevention Network \\
\hline

Hot Spring Tourism & Water Quality Management & Mineral Balance Analysis Report & Geological Origin Display \\
 &  & Real-Time Monitoring and Warning System & Temperature Stability Control \\
 & Health Intervention & TCM Constitution Identification Process & Stress Hormone Regulation Technology \\
 &  & Sleep Quality Improvement Therapy & Multimodal Healing Design \\
 & Environmental Design & Privacy Level Standards & Cultural Landscape Integration \\
 &  & Sound and Light Environment Adaptation & Traditional Wellness Method Integration \\
\hline

Ice and Snow Tourism & Snowfield Engineering & International Certified Slope Planning & Snow Compaction Operational Standards \\
 &  & Buffer Zone Safety Design Standards & Avalanche Warning and Response System \\
 & Skill Training & Standardized Movement Teaching System & Turn Technique Decomposition Course \\
 &  & Speed Control Training Module & Injury Prevention Education \\
 & Equipment Service & Board Length Adaptation Algorithm & Maintenance Knowledge Base \\
 &  & Disinfection Process Visualization & Rental Cost-Effectiveness Analysis Model \\
\hline

Island Tourism & Ecological Restoration & Coral Bleaching Warning Mechanism & Coral Adoption Survival Tracking \\
 &  & Gene Bank Diversity Protection & Marine Debris Management Initiative \\
 & Activity Safety & Snorkeling Window Prediction Model & Satellite Tracking Data Application \\
 &  & Dynamic Electronic Fence Deployment & Extreme Weather Response Plan \\
 & Cultural Inheritance & Island Handicraft Heritage Genealogy & Dialect Preservation and Recording \\
 &  & Local Cuisine Certification System & Immersive Maritime History Exhibition \\
\hline

Religious Tourism & Taboo Management & Cross-Religious Conflict Warning System & Electronic Ticket Anti-Counterfeiting \\
 &  & Graded Access Permit Process & Holy Site Reservation Response Timeliness \\
 & Artifact Preservation & Ancient Architecture Restoration Standards & Ancient Manuscript Digital Archiving \\
 &  & Microenvironment Control Technology & Incense Fire Safety Monitoring System \\
 & Virtual Experience & Blockchain Merit Ledger System & AR Architectural Structure Analysis \\
 &  & Metaverse Ritual Scene Restoration & Multilingual Scripture Intelligent Interpretation \\
\hline

Urban Tourism & Spatial Intelligence & Underground Navigation Path Optimization & Real-Time Crime Heatmap Monitoring \\
 &  & Digital Twin City Modeling & Microclimate Dynamic Adjustment \\
 & Memory Inheritance & Oral History Recording and Transcription & Industrial Heritage Revitalization \\
 &  & Streetscape Temporal Comparison System & Immigrant Culture Thematic Display \\
 & Update Assessment & Gentrification Impact Control Strategy & Smart Facility Operation Standards \\
 &  & Authenticity Protection Metrics & Community Interest Negotiation Mechanism \\
\midrule
Characteristic Town Tourism & Authenticity Preservation & Traditional Shop Revitalization Ratio & Building Age Detection Technology \\
 &  & Ancient Production Process Restoration & Dialect Usage Frequency Statistics \\
 & Military Heritage & Defensive Fortification Scene Restoration & Fortress Attack-Defense Simulation System \\
 &  & Cold Weapon Martial Arts Experience & Wall Settlement Monitoring Frequency \\
 & Craft Revitalization & Intangible Heritage Genealogy Compilation & Metaverse Course Development \\
 &  & Raw Material Gene Protection Plan & Handicraft Consumption Scene Design \\
 \hline
 
Family Study Tourism & Ability Adaptation & Zone of Proximal Development Design & Learning Style Identification Tool \\
 &  & Multiple Intelligences Matching Model & Special Needs Children Support Plan \\
 & Cognitive Construction & Interdisciplinary Teaching Module & Scientific Experiment Error Control Standards \\
 &  & Metacognitive Training Frequency & Reflective Journal Evaluation System \\
 & Scene Restoration & Historical Tool Reproduction Accuracy & Multisensory Stimulation Parameter Design \\
 &  & Scientific Phenomenon Simulation Experiment & Safety Education Integration Plan \\
\hline

Wellness Tourism & Healing Environment & Forest Volatile Organic Compound Monitoring & Negative Ion Dynamic Regulation \\
 &  & Hot Spring Mineral Penetration Technology & Five-Sense Balance Adjustment Plan \\
 & Health Management & Personalized Therapy Phase Design & Biological Rhythm Synchronization Technology \\
 &  & Precise Nutritional Meal Allocation & Chronic Disease Intervention Pathway \\
 & Risk Control & Allergen Isolation Database & Equipment Safety Certification Standards \\
 &  & Emergency Network Coverage & Emergency Plan Drill Frequency \\
\hline

 &  & Flow Experience Matching Logic & Dynamic Scene Switching Algorithm \\
 & Intelligent Service & User Profile Matching Recommendation & Accessibility Facility Coverage \\
 &  & Real-Time Demand Response Timeliness & Multilingual Support Level System \\  
 & Environmental Control & Noise Zoning Management Standards & Queue Optimization Algorithm Logic \\
 &  & Temperature-Humidity-Wind Precision Regulation & Cultural IP Derivative Development Path \\
\hline

\end{tabular}
\captionsetup{font=scriptsize}
\caption{Tourism Characteristic ToT Elements Framework (partial). This table presents a ToT framework, encompassing tourism categories (\textbf{Category}), each comprising specific elements (\textbf{Element}), with every element further subdivided into detailed sub-elements.}
\label{tab:framework}
\end{table}

\end{onecolumn}

\end{document}